\documentclass[letterpaper, 10 pt, conference]{ieeeconf}  

\usepackage{graphicx}
\usepackage{subcaption}
\usepackage{diagbox}
\usepackage{float} 
\usepackage{caption}
\usepackage{lipsum}
\usepackage{multirow} 
\usepackage{amsmath}
\usepackage{algorithm}
\usepackage{algpseudocode}
\usepackage{stfloats}
\usepackage{placeins}
\usepackage{amsfonts, amssymb}
\usepackage{microtype}
\usepackage{url}
\usepackage[percent]{overpic}
\usepackage{xcolor} 
\usepackage{tikz}
\usepackage{tabularx}
\usepackage{placeins}
\usepackage{amsmath}  
\usepackage[letterpaper, top=72pt, bottom=54pt, left=54pt, right=54pt]{geometry}  

\IEEEoverridecommandlockouts  

\title{\LARGE \bf
Tethered Multi-Robot Systems in Marine Environments
}

\author{Markus Buchholz$^{1}$, Ignacio Carlucho$^{1}$, Michele Grimaldi$^{1}$ and Yvan R. Petillot$^{1}$ %
\thanks{$^{1}$ School of Engineering \& Physical Sciences, Heriot-Watt University, Edinburgh, UK
        {\tt\small m.buchholz@hw.ac.uk}}%
}

\begin{document}

\maketitle
\thispagestyle{empty}
\pagestyle{empty}

\begin{abstract}
This paper introduces a novel simulation framework for evaluating motion control in tethered multi-robot systems within dynamic marine environments. Specifically, it focuses on the coordinated operation of an Autonomous Underwater Vehicle (AUV) and an Autonomous Surface Vehicle (ASV). The framework leverages GazeboSim, enhanced with realistic marine environment plugins and ArduPilot's Software-In-The-Loop (SITL) mode, to provide a high-fidelity simulation platform. A detailed tether model, combining catenary equations and physical simulation, is integrated to accurately represent the dynamic interactions between the vehicles and the environment. This setup facilitates the development and testing of advanced control strategies under realistic conditions, demonstrating the framework's capability to analyze complex tether interactions and their impact on system performance.\footnote{\url{https://github.com/markusbuchholz/marine-robotics-sim-framework}} 
\end{abstract}


\section{INTRODUCTION}

Recent advances in marine robotics have catalyzed the development of integrated multi-robot systems for demanding offshore applications such as wind farm inspections, environmental monitoring, and subsea infrastructure maintenance \cite{LIU2022112187, ZEREIK2018350}. A prevalent approach in these environments involves the use of Autonomous Underwater Vehicles (AUVs) and Autonomous Surface Vehicles (ASVs) interconnected by a flexible tether, which extends mission endurance while ensuring robust communication and power supply.

The dynamic behavior of the tether introduces unique experimental challenges and opportunities. Functioning as a flexible joint, the tether interacts with environmental disturbances—such as waves, ocean currents, and wind—and influences the overall motion of the vehicles \cite{battocletti2024entanglement, Rajan_Nagendran_2016}. These interactions, if not carefully studied, can lead to unforeseen operational issues, including entanglement and unanticipated vehicle dynamics.

\begin{figure}[t]
\centering
\includegraphics[width=0.47\textwidth]{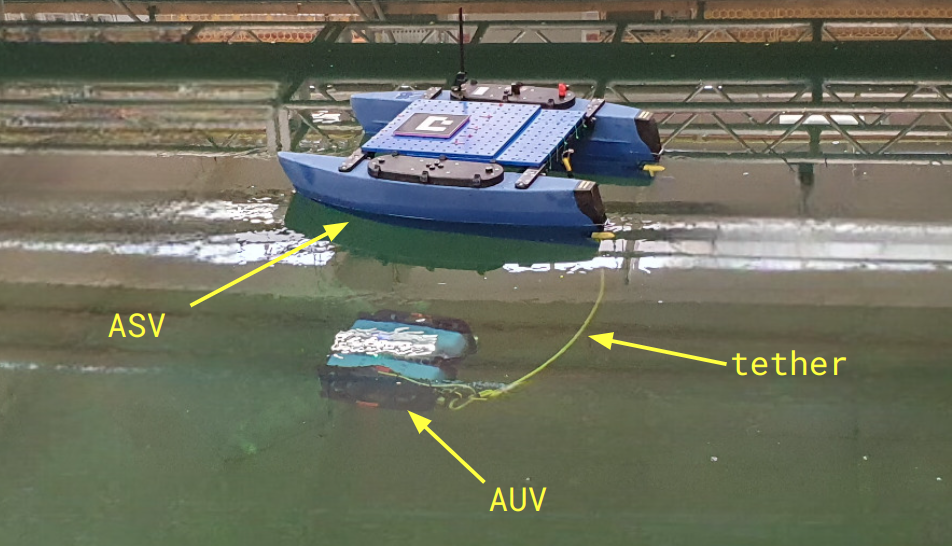}
\caption{Multi-robot tethered system simulated in this paper.}
\label{fig:intro}
\end{figure}

To address these complexities, our study introduces a novel experimental framework implemented in GazeboSim \cite{gazebosim}. This high-fidelity simulation platform provides a one-to-one replication of real-world operations, offering a unique capability to experiment with tether dynamics and its interactions with both vehicles and environmental disturbances. The framework uniquely integrates: \begin{itemize} \item \textbf{Realistic Tether Simulation:} A comprehensive open-source model of the tether as a flexible joint, capturing its response to environmental forces and vehicle motions. \item \textbf{Multi-Vehicle Coordination:} An experimental setup that facilitates the study of cooperative interactions between AUVs and ASVs, with an emphasis on the influence of tether dynamics rather than control optimization. \item \textbf{Environmental Interaction:} The ability to simulate realistic marine disturbances—such as waves, currents, and wind—to evaluate their effects on tether behavior and vehicle performance. \end{itemize}

A noteworthy aspect of our approach is the use of Ardupilot’s (SITL) mode within the simulator, which closely replicates the behavior of actual vehicles. This replica simulation, complemented by real-world tests, ensures that the operational dynamics observed in GazeboSim are representative of practical deployments.

By delivering this comprehensive and open-source experimental framework, our work provides researchers with a powerful tool to rigorously test and analyze the interplay between tether dynamics and multi-vehicle operations under realistic marine conditions. The seamless integration of replica simulation with real-world testing enhances the credibility and applicability of the insights gained, thereby advancing the state-of-the-art in marine robotics research.

In the following sections, we detail the design and implementation of our framework, demonstrating its capability to simulate complex tether interactions and their impact on system performance in dynamic ocean environments.

\begin{figure*}[tp]
    \centering
    \includegraphics[width=.65\textwidth, height=0.45\textheight]{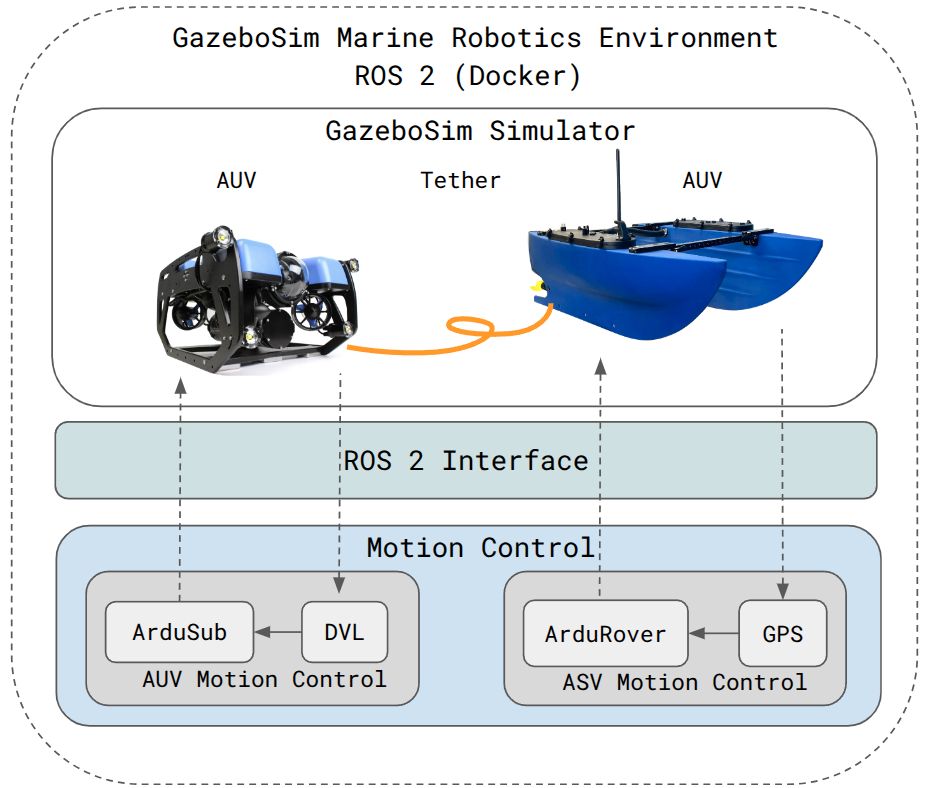}
    \caption{System overview of a tethered ASV-AUV collaboration framework for autonomous navigation.}
    \label{fig:img1}
\end{figure*}

\section{RELATED WORK}

The development of robust and realistic simulation frameworks is crucial for advancing marine robotics, particularly for complex systems such as tethered ASV–AUV collaborations \cite{smith2019marine, johnson2020dynamic}. These frameworks must accurately model the intricate dynamics of marine environments, including hydrodynamic forces, tether behavior, and multi-robot interactions \cite{johnson2020dynamic}.

Several simulation platforms exist with varying capabilities. For instance, Stonefish \cite{stonefish},HoloOcean \cite{holoocean} platform—offers high-fidelity visual rendering and physics simulation for underwater environments. However, despite its visual advantages, Stonefish does not natively support widely used control software such as ArduPilot, which complicates the direct translation of simulation results to real-world applications \cite{lee2018challenges}.

Conversely, GazeboSim provides a versatile, open-source simulation environment that supports robotic applications in marine settings \cite{gazebosim}. Its extensibility allows for the integration of diverse physics engines and sensor models \cite{martinez2017simulation}. Nonetheless, achieving realistic underwater dynamics and accurate tether modeling in GazeboSim requires significant customization \cite{wang2016underwater}.

Recent research has advanced the realism of underwater simulations by incorporating detailed hydrodynamic models and environmental effects. Fossen et al. \cite{fossen2011handbook} provide a comprehensive theoretical foundation for marine vehicle dynamics and control, which underpins accurate simulation modeling. Furthermore, several studies have focused on modeling flexible tethers—using methods such as catenary equations, finite element analysis, and dynamic modeling—to capture cable behavior in underwater environments \cite{tether1, Tortorici2024, tether2, gupta2019tether}. These models are essential for representing the forces and constraints imposed by tethers in multi-robot systems.

In the realm of multi-robot systems, simulation frameworks are indispensable for prototyping and validating algorithms in path planning, motion control, and cooperative behaviors. The Robot Operating System (ROS) has emerged as a standard platform for integrating robotic software components, thereby facilitating complex multi-robot simulations \cite{ros}. Integrating ROS with simulators like GazeboSim enables the testing of advanced control strategies in realistic settings \cite{brown2018integrated}. Recent work on bridging the MAVLink protocol with ROS\,2 has further enhanced communication between simulated and real robotic platforms \cite{rosMavlink}.
Our work builds on these developments by providing an open-source framework that integrates GazeboSim with ArduPilot’s (SITL) mode, achieving high-fidelity simulations of tethered ASV–AUV systems. By leveraging realistic vehicle dynamics and incorporating detailed tether modeling, our framework serves as a powerful testbed for prototyping and evaluating multi-robot coordination algorithms in dynamic marine environments. Moreover, the direct support for ArduPilot within our framework ensures a seamless transition from simulation to real-world deployment, thereby enhancing system reliability and safety \cite{park2019field}.
Furthermore, our framework simulates environmental disturbances such as ocean currents and waves, which are critical for assessing the robustness of control algorithms under realistic conditions \cite{wang2016underwater}. This capability is particularly vital for tethered systems, where environmental forces can significantly impact tether behavior and overall system performance.

\section{SYSTEM ARCHITECTURE}

This work presents a simulation framework for tethered marine robotic systems built on ROS\,2 within a Docker container, providing a robust and portable environment for multi-robot experimentation. Fig.~\ref{fig:img1} depicts the overall architecture, where GazeboSim serves as the high-fidelity simulator. GazeboSim is enhanced with specialized plugins to model realistic marine phenomena, including waves, ocean currents, wind, and hydrodynamic interactions.

The framework integrates two primary vehicles:
\begin{enumerate}
    \item An AUV, represented by a BlueROV2, which employs ArduSub for motion control and a Doppler Velocity Log (DVL) for precise underwater navigation.
    \item An ASV, modeled using ArduRover and equipped with a GPS sensor for accurate surface localization.
\end{enumerate}
These vehicles are interconnected by a simulated tether that accurately captures catenary effects and environmental forces.

To ensure realistic operation, the framework incorporates SITL by running simultaneous instances of ArduSub and ArduRover, (both branches of ArduPilot), enabling authentic control and sensor feedback via the MAVLink protocol. A dedicated bridging interface translates MAVLink messages to native ROS\,2 topics, providing seamless data exchange between the autopilot systems and the rest of the simulation. The integrated Motion Control layer within GazeboSim supports concurrent autopilot operation, while the ROS\,2 environment offers a modular structure for implementing and evaluating advanced control strategies. By combining Docker-based portability, ROS\,2 modularity, SITL fidelity, and comprehensive modeling of tether dynamics and environmental effects, this architecture serves as a powerful testbed for analyzing the complex interactions between the AUV, ASV, and the marine environment. Researchers can thus develop, validate, and refine advanced tethered control algorithms under realistic operational conditions.

\section{Dynamic modeling}

\subsection{Modeling of ASV and AUV}

The dynamics of marine vehicles, such as ASV or AUV, are typically modeled using rigid body equations of motion in planar motion \cite{fossen2011handbook}:
\begin{equation}
\mathbf{M} \dot{\mathbf{\upsilon}} + \mathbf{C}(\mathbf{\upsilon}) \mathbf{\upsilon} + \mathbf{D}(\mathbf{\upsilon}) \mathbf{\upsilon} + g(\eta)= \mathbf{T} + \Delta,
\end{equation}
where $\mathbf{\upsilon}$ is the body frame velocity vector, $\mathbf{M}$ is the mass and inertia matrix, $\mathbf{T}$ is the control input vector, $\mathbf{C}$ is the Coriolis and centrifugal forces matrix, $\mathbf{D}$ is the damping matrix, $g(\eta)$ are the buoyancy elements, and $\Delta$ represents environmental disturbances. These disturbances, including wave forces, wind forces, and ocean currents \cite{dist_1, dist_2, dist_6}, are modeled as follows:

\noindent\textbf{Wave Forces:} Calculated based on water density \(\rho_{\text{water}}\), gravity \(g\), ASV dimensions (beam \(B\), length \(L\), and draft \(T\)), wave amplitude \((A)\), wavelength \((\Lambda)\), and wave parameters. Specifically,
\begin{equation}\label{eq:wave}
F_{\text{wave}} 
= \rho_{\text{water}} \cdot g \cdot B \cdot L \cdot T \cdot k,
\end{equation}
where 
\[
k = \biggl(\frac{2 \pi}{\Lambda}\biggr)\,A
\]
is a constant representing wave characteristics.

\noindent\textbf{Wind Forces:} Modeled based on air density \(\rho_{\text{air}}\), wind speed \(V_w\), drag coefficients \((C_x, C_y)\), and projected area for wind force \((A_w, A_{lw})\):
\begin{equation}\label{eq:wind}
F_{\text{wind}} 
= 0.5 \cdot \rho_{\text{air}} \cdot V_w^2 \cdot C_x \cdot A_w.
\end{equation}

\noindent\textbf{Current Effects:} Represented by current speed \((V_c)\) and direction angles \((\alpha_c, \beta_c)\) affecting ASV motion:
\begin{equation}\label{eq:current}
v_c = V_c \,\cos(\alpha_c)\,\cos(\beta_c).
\end{equation}

\subsection{Tether modeling}

\begin{figure}[t]
\centering
\includegraphics[width=0.47\textwidth, height=0.3\textheight]{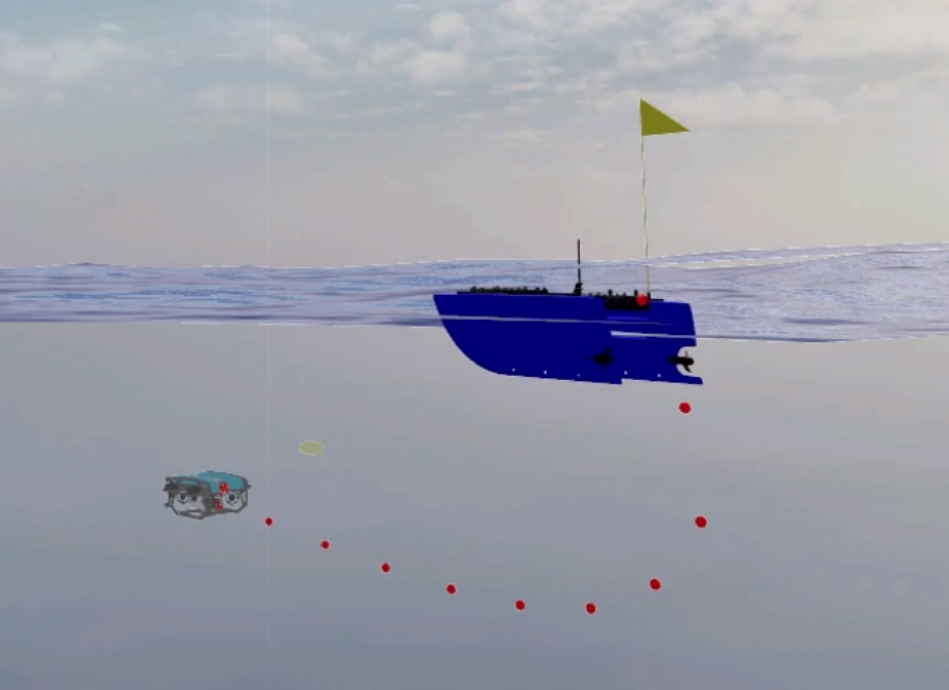}
\caption{Schematic of the tether model showing the ASV, a series of buoy segments connected by ball joints, and the AUV. The joints enforce angular limits and include damping to simulate realistic tether behavior.}
\label{fig:tether_model}
\end{figure}

In tethered marine systems involving an ASV and an AUV, the modeling of the tether plays a critical role. We introduce a novel approach by integrating a comprehensive tether model into the GazeboSim environment, enabling high-fidelity simulations of ASV-AUV coordination.
Our tether model combines an analytical catenary formulation to describe the cable shape with a 
detailed representation of the physical properties of the tether and its connection via joints, 
all implemented within GazeboSim. The idealized shape of the hanging cable is given by the 
\emph{catenary} curve. For our simulation, we assume the cable is inextensible and flexible. 
The standard catenary equation is:
\begin{equation}\label{eq:catenary_standard}
z(x) = a \cosh\!\bigl(\tfrac{x}{a}\bigr),
\end{equation}
which is modified to satisfy the boundary conditions at the ASV and AUV:
\begin{equation}\label{eq:catenary_modified}
z(x) = a \cosh\!\bigl(\tfrac{x - c}{a}\bigr) \;-\; a \cosh\!\bigl(\tfrac{c}{a}\bigr).
\end{equation}
The parameters \(a\) and \(c\) are determined by solving the following constraints numerically:

\noindent\textbf{Vertical Displacement:}
\begin{equation}
a \Bigl[\cosh\!\bigl(\tfrac{D-c}{a}\bigr) \;-\; \cosh\!\bigl(\tfrac{c}{a}\bigr)\Bigr] = \Delta z,
\end{equation}
where \(D\) is the horizontal distance and \(\Delta z = z_{\text{AUV}} - z_{\text{ASV}}\).
\noindent\textbf{Total Cable Length:}
\begin{equation}
L = a \Bigl[\sinh\!\bigl(\tfrac{D-c}{a}\bigr) \;+\; \sinh\!\bigl(\tfrac{c}{a}\bigr)\Bigr],
\end{equation}
where \(L\) is the total length of the tether. In GazeboSim, the tether is modeled as a series of buoyant spheres (or buoy segments) connected by ball joints, simulating real-world flexibility. Each segment has defined mass, inertia (calculated based on geometry), and near-neutral buoyancy. The ball joints between segments allow rotation in all directions and are configured with angular limits (e.g., \([-1.5, 1.5]\) radians) and damping (e.g., \(\text{joint\_damping} = 0.05\)) to ensure realistic behavior and simulation stability. A schematic of this model is shown in Fig.~\ref{fig:tether_model}.

The integration of this detailed tether model into GazeboSim offers significant advantages. It allows for the simulation of cable flexibility, the dynamic effects of gravitational forces, buoyancy, hydrodynamic loads, and joint damping. GazeboSim's physics engine handles these interactions, providing a realistic tether response to environmental forces and vehicle motions. The analytical catenary model provides a static shape approximation, while the physical simulation captures dynamic complexities. This dual approach, enabled by the GazeboSim integration, represents a substantial advancement in our capability to accurately model and simulate tethered marine systems, providing valuable insights for the development and control of ASV-AUV collaborative missions.

The presented tether model, which combines a classical catenary formulation with a detailed physical simulation within the GazeboSim environment, offers a comprehensive and realistic approach to modeling tethered ASV-AUV systems. This integration facilitates a deeper understanding of the tether's behavior under various conditions, contributing significantly to the advancement of marine robotics research and applications.


\begin{figure}[tp]
    \centering
    \includegraphics[width=0.5\textwidth, height=0.25\textheight]{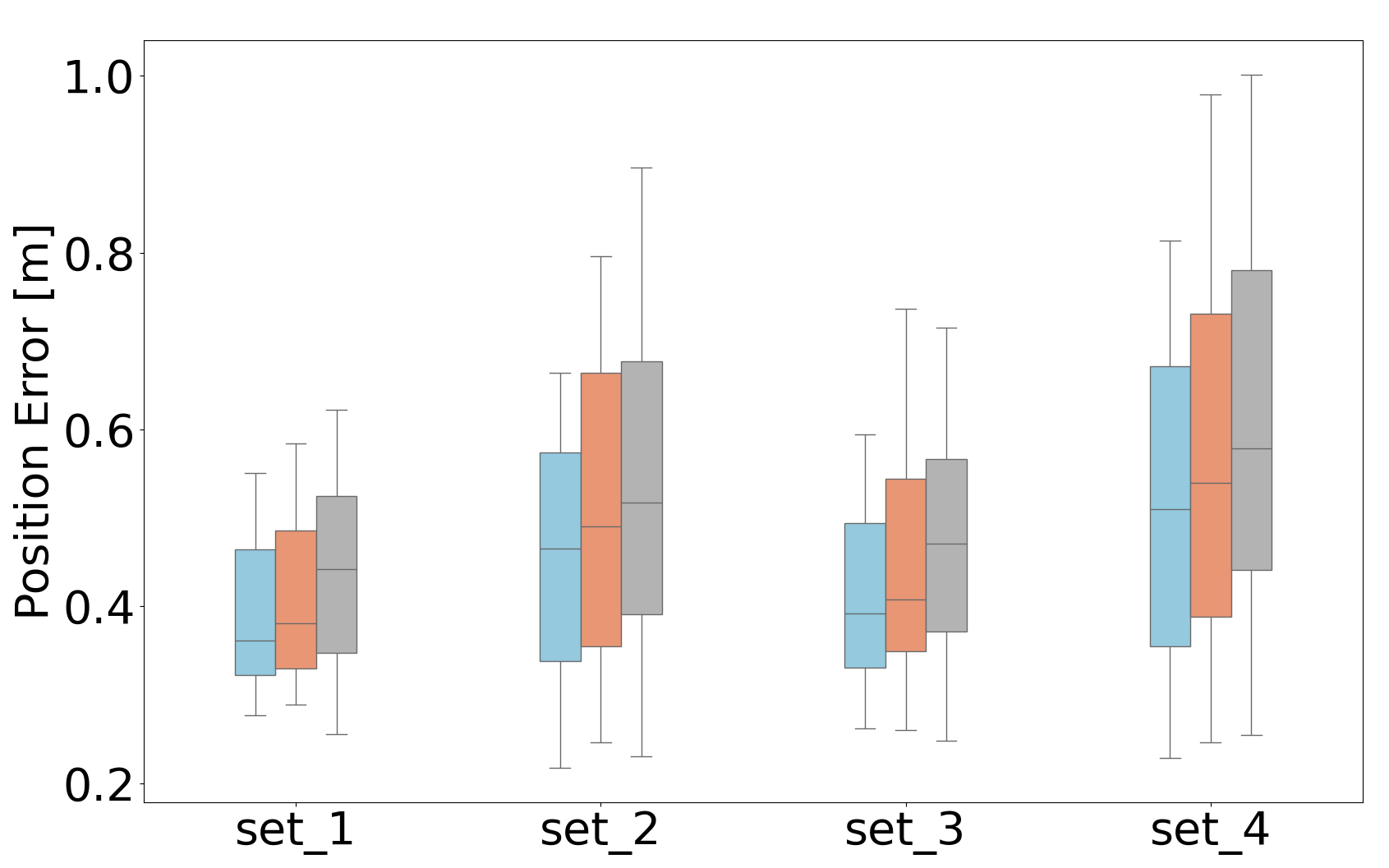}
    \caption{Errors for ASV and AUV at different tether lengths: 5m (blue), 10m (orange), and 15m (grey) under rough sea conditions.}
    \label{fig:scenario_tether}
\end{figure}


\section{EVALUATION}

Our research pipeline heavily leveraged GazeboSim to evaluate the performance of our proposed methodologies for tethered ASV-AUV systems. GazeboSim's ability to accurately simulate marine environments, coupled with its integration with ROS 2, provided a realistic platform for testing and validating our algorithms.

For our experiments, we utilized the BlueBoat (ASV) \cite{bluerobotics} and the BlueROV2 Heavy (AUV) \cite{bluerobotics}, simulated within GazeboSim. These platforms were configured to match their physical counterparts, ensuring realistic simulations. The experiments were conducted using GazeboSim with marine robotics plugins \cite{mainwaring_asv_wave_sim}, replicating offshore conditions. ROS 2 was used for real-time motion control, testing the ASV-AUV interactions under varying environmental disturbances. We designed scenarios to mimic realistic offshore operations, including navigation around underwater structures.

Simulations were performed under Moderate (3 m/s wave height, 3 m/s wind, 0.5 m/s currents) and Rough (4.5 m/s wave height, 4.5 m/s wind, 1.0 m/s currents) sea conditions to assess system robustness.

\subsection{Simulation Results and GazeboSim's Role in Methodology Evaluation}

GazeboSim's visualization capabilities were crucial for understanding system behavior. It allowed us to observe the ASV and AUV paths and tether dynamics in real-time, enabling analysis of synchronization under environmental disturbances. This visual feedback aided in identifying areas for improvement.

We conducted comparative studies under different environmental conditions and system configurations. As shown in Fig. \ref{fig:scenario_tether}, we evaluated motion performance under Moderate and Rough seas, observing the impact of disturbances on accuracy. Testing four configurations, including disturbance-aware and non-aware modes, highlighted the importance of disturbance compensation for improved performance. The results indicate that configurations with disturbance compensation (set\_1 and set\_3) consistently showed lower position errors compared to those without (set\_2 and set\_4), especially under rough sea conditions.

Specifically, we tested undisclosed methods focusing on tether length impact. GazeboSim facilitated the evaluation of combined position errors for varying tether lengths (5m, 10m, 15m) under rough seas. The results, as depicted in Fig. \ref{fig:scenario_tether}, show that longer tethers generally lead to increased position errors. This can be attributed to the increased freedom and exposure to environmental disturbances with longer tethers.

In summary, GazeboSim provided a realistic and versatile platform for evaluating our methodologies. Its simulation accuracy and visualization capabilities allowed us to validate our algorithms and gain insights into tethered system behavior under various conditions.

\section{CONCLUSION}

This paper presented a comprehensive simulation framework for tethered ASV-AUV systems, emphasizing the use of GazeboSim to evaluate motion control strategies in dynamic marine environments. The framework integrates detailed tether modeling with realistic environmental simulations, providing a robust platform for testing and validating advanced control algorithms. The evaluation, conducted through various simulation scenarios, demonstrated the framework's capability to accurately represent complex interactions and its potential for advancing the development of tethered multi-robot systems. Future work will focus on extending the framework's applicability to real-world platforms and integrating additional functionalities, such as robotic arm manipulation, to further enhance its utility in offshore operations.

\section{ACKNOWLEDGMENTS}
This research was supported by the EPSRC project Underwater Intervention for offshore renewable energies (UNITE) under the Grant Reference EP/X024806/1.
\FloatBarrier
\bibliographystyle{IEEEtran}
\bibliography{reference}
\end{document}